\pdfoutput=1

\documentclass[11pt]{article}

\usepackage[final]{acl}
\usepackage{booktabs}
\usepackage{times}
\usepackage{adjustbox}

\usepackage{latexsym}
\usepackage{numprint} 
\usepackage{pgf-pie}
\usepackage{pgfplots}
\pgfplotsset{compat=1.17}
\usepackage{pgfplotstable}
\usepackage{filecontents}
\usepackage[T1]{fontenc}

\usepackage[utf8]{inputenc}

\usepackage{microtype}
\usepackage{float}
\usepackage{tikz}
\usetikzlibrary{shapes.geometric, arrows.meta, positioning, backgrounds, fit, calc}
\usepackage{fontawesome}
\tikzstyle{input}  = [rectangle, rounded corners, draw=blue!60, fill=blue!10, thick, minimum height=1cm, minimum width=2.5cm]
\tikzstyle{icon}   = [circle, draw=gray!60, fill=gray!20, thick, minimum size=1cm]
\tikzstyle{output} = [rectangle, rounded corners, draw=olive!60, fill=olive!10, thick, minimum height=1cm, minimum width=3cm]
\tikzstyle{arrow}  = [-{Latex[length=3mm,width=2mm]}, thick]

\usepackage{inconsolata}

\usepackage[table]{xcolor}

\usepackage{graphicx}

\title{A Simple Data Augmentation Strategy for Text-in-Image Scientific VQA}

\author{
    Belal Shoer, Yova Kementchedjhieva \\
    MBZUAI \\
    \texttt{\{belal.shoer,yova.kementchedjhieva\}@mbzuai.ac.ae}
}

\begin{document}
\maketitle
\begin{abstract}
Scientific visual question answering poses significant challenges for vision-language models due to the complexity of scientific figures and their multimodal context. Traditional approaches treat the figure and accompanying text (e.g., questions and answer options) as separate inputs. EXAMS-V introduced a new paradigm by embedding both visual and textual content into a single image. However, even state-of-the-art proprietary models perform poorly on this setup in zero-shot settings, underscoring the need for task-specific fine-tuning. To address the scarcity of training data in this "text-in-image" format, we synthesize a new dataset by converting existing separate image-text pairs into unified images. Fine-tuning a small multilingual multimodal model on a mix of our synthetic data and EXAMS-V yields notable gains across 13 languages, demonstrating strong average improvements and cross-lingual transfer.\footnote{Dataset: \url{https://huggingface.co/datasets/Shoir/Scientific_VQA}}

\end{abstract}

\section{Introduction}
Vision-language models (VLMs) have advanced AI by enabling multimodal reasoning, facilitating more natural user interaction in tasks such as Visual Question Answering (VQA) and captioning. \citet{Antol_2015_ICCV} proposed VQA as a task that spans language and image to generate an accurate response. The VQA task has evolved rapidly with applications and benchmarks in domains such as science \citep{lu2022learn}, chart understanding \citep{masry-etal-2022-chartqa}, document analysis \citep{DBLP:journals/corr/abs-2007-00398}, medical imaging \citep{ImageCLEFVQA-Med2018}, and other real-world applications. VQA tasks typically follow either a multiple-choice or open-ended format.

\begin{figure}[!ht]
  \centering
  \resizebox{\columnwidth}{!}{%
    \begin{tikzpicture}[node distance=0.6cm and 1cm, every node/.style={inner sep=0pt}]
      
      \node[
  draw,
  fill=blue!5,
  rounded corners,
  inner sep=6pt 
] (fig) {
  \includegraphics[width=5cm]{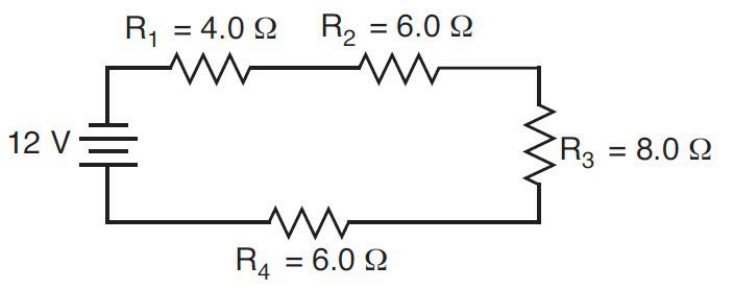}
};

  \node[
  draw,
  fill=blue!5,
  rounded corners,
  right=2cm of fig,
  minimum width=5.5cm,
  minimum height=4cm,
  inner sep=6pt,
  align=left
] (text) {
  \parbox{5.26cm}{%
    \large
    The circuit diagram below represents four resistors connected to a 12-volt source.

    \vspace{0.5em}

    What is the total current in the circuit?

    \vspace{0.5em}

    (1) 0.50 A \quad (2) 2.0 A \\
    (3) 8.6 A \quad \hspace{0.5em}(4) 24 A
  }
};

      \node[align=center, above=2.3cm of $(fig)!0.5!(text)$, font=\bfseries] {Before: Disjoint Image and Text};

      \node[font=\bfseries\LARGE, align=center] at ($(fig.east)!0.5!(text.west)$) {+};

      \coordinate (arrowstart) at ($(fig.south)!0.5!(text.south)$);
\coordinate (arrowend) at ($(arrowstart) + (0,-0.8)$);
\draw[->, thick] (arrowstart) -- (arrowend);

      \node[draw=green!60!black, fill=green!5, rounded corners, below=1cm of arrowend, inner sep=6pt] (unified) {
  \begin{minipage}[c][5.5cm][c]{9cm}
    \centering
    \includegraphics[width=\columnwidth, keepaspectratio]{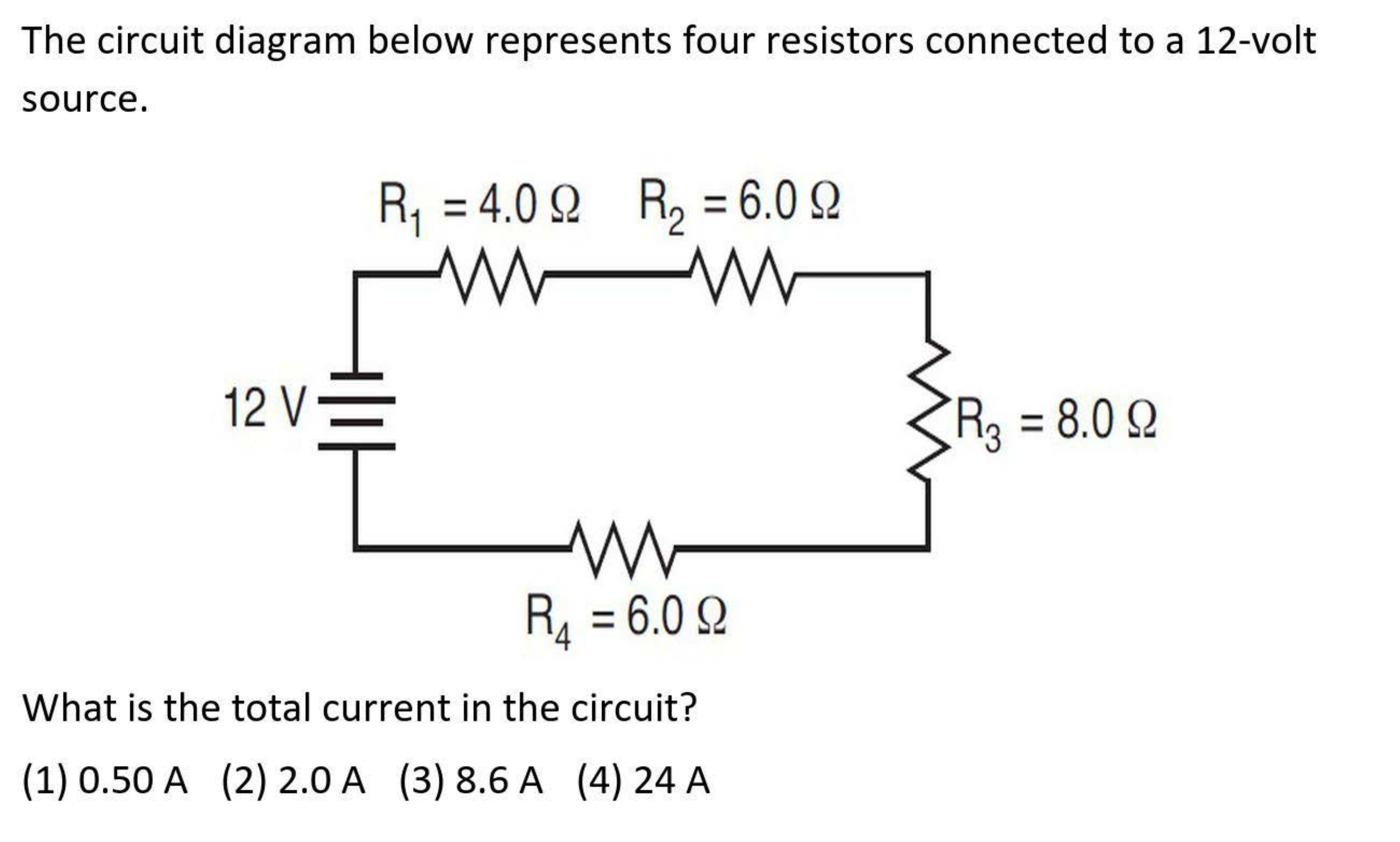}
  \end{minipage}
};

      \node[align=center, above=0.2cm of unified, font=\bfseries] {After: Unified Image + Question};

    \end{tikzpicture}
  }
  \captionsetup{skip=1em}
  \caption{Synthetic data generation via mapping of a disjoint figure and text into a unified image.}
  \label{fig:before_after_blocks}
\end{figure}
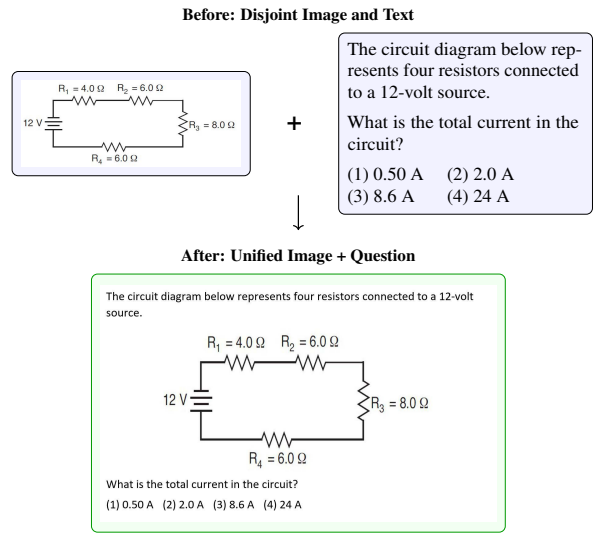

In multiple-choice scientific VQA, the input typically consists of an image (figure, table, chart) accompanied by a question and answer choices in text form. The task requires reasoning over both image and text components to select the correct answer. These modalities are often processed separately in multimodal models. However, in practice, the text is often embedded within the visual modality, for example, screenshots of digital exams or textbook photos.
To address this, \citet{das2024exams} introduced a new scientific VQA benchmark that consists of images with embedded questions, providing a robust benchmark for evaluating model performance under realistic conditions. The EXAMS-V dataset includes two splits: train (16.5K instances) and test (4.8K instances), spanning 15 languages.

This text-in-image formulation of scientific VQA either requires a separate Optical Character Recognition (OCR) step, which may introduce noise, or, preferably, VLMs with strong inherent OCR capabilities that can jointly reason over visual content and embedded text.
Yet, current VLMs typically benefit from text as opposed to text-in-image format, even if the text is just describing the contents of the image itself \citep{vineet2024is}.    
EXAMS-V approached the task in a zero-shot manner, without leveraging the training data. Fine-tuning for reasoning over visually-embedded text is likely to improve performance.

While we already have the EXAMS-V training split in the text-in-image format, we find that it provides limited coverage, with an average of 1,415 data points per language. To address this, we augment the training set by synthesizing text-in-image data points derived from disjoint scientific VQA datasets, in four languages: Chinese, English, Italian, and German. This results in approximately 1,742 additional examples on average per language.

\section{Background}
In this section, we provide the necessary background for our work. We begin by reviewing traditional datasets and benchmarks commonly used in scientific VQA, highlighting their strengths and limitations. We then review EXAMS-V, the primary benchmark our work builds upon, and describe our chosen VLM, PaLIGemma \citep{steiner2024paligemma}, explaining the rationale behind its selection. 
\subsection{Traditional Datasets}
There are a number of multi-modal scientific VQA datasets that span multiple scientific fields, such as physics, chemistry, biology, mathematics, and geology. ScienceQA \citep{lu2022learn}, was introduced as an English monolingual scientific multimodal dataset that has been collected from elementary and high school curricula. MMMU \citep{yue2023mmmu} is another English-language scientific benchmark compiled from college exams and textbooks to challenge the VLMs' abilities on multi-modal multi-discipline subjects. These datasets treat vision and language as separate inputs, whereas EXAMS-V presents a novel approach by combining both modalities in a single image.

We harvest our data from 5 different datasets namely, M3EXAM \citep{zhang2023m3exam}, CMMU \citep{he2024cmmu}, M4U \citep{wang2024m4u}, MMMU-PRO \citep{yue2024mmmu}, and Pinocchio \citep{pinocchio2024nlu}. Since these datasets separate language and vision components, we synthetically combine the question and answer text with the corresponding figures to create text-in-image training examples.

Synthetic data generation has been shown to improve VLMs’ performance. \citet{chen2024allava} generated a dataset of 1.3M examples and showed that small models can match or even outperform larger ones when trained on synthetic data. Moreover, \citet{liu2024synthvlm} reported improved performance using their synthetic dataset.

\subsection{Text-in-Image Datasets: EXAMS-V}
The composition of language and vision poses a significant challenge to VLMs. \citet{wang2024picture} found that in spatial reasoning tasks, VLMs rarely outperform their traditional LLM counterparts and when provided with both image and text, they rely less on the visual modality.

EXAMS-V is a multilingual multimodal benchmark that consists of 20,932 multiple-choice questions curated from national exams from multiple nations. It contains two data formats: 15,846 text-only and 5,086 text-and-visual images.

\subsection{PaliGemma}
VLMs are widely adopted for their strong generalization across tasks such as image captioning, VQA, and visual grounding. Google recently introduced PaLIGemma 2, an enhanced version of PaLIGemma that integrates the more powerful Gemma 2 language model together with the SigLIP vision encoder. It supports three image resolutions: 224$^2$, 448$^2$, and 896$^2$ pixels.

PaLIGemma 2 was trained in stages: initially on 1 billion image-text pairs at 224$^2$ using the combined SigLIP So400m and Gemma 2 checkpoints, followed by 50 million examples at 448$^2$ and 10 million at 896$^2$, and finally on a mix of academic tasks including VQA, captioning, and detection.

We chose to fine-tune PaLIGemma 2 for three main reasons. First, it is lightweight, making it a practical alternative to large proprietary models. Second, it supports 34 languages, aligning well with our multilingual goals. Third, it offers flexibility in size and resolution and has been pre-trained on tasks relevant to our setting.

\section{Data Augmentation}

This section outlines our method for generating synthetic text-in-image instances for VQA and introduces the pre-trained VLM used in our experiments. We focus on the text-with-visual format rather than the text-only due to its limited presence in EXAMS-V, with only 5,162 such images.

We use data from five datasets that provide separate text and image pairs: M3EXAM, CMMU, M4U, MMMU-PRO, and Pinocchio. These datasets span multiple languages and subjects. We focus on Chinese, English, Italian, and German. We filter the data to retain only science-related instances, primarily from Chemistry, Physics, Biology, Biochemistry, and Engineering. Each instance is formatted consistently, with the question at the top, followed by the figure and answer options. For an example of our method, refer to Figure~\ref{fig:before_after_blocks}.

To simulate realistic exam formats, Hanzi and Latin scripts are rendered using randomly selected fonts and dark text colors. We use common fonts such as SimSun and SimHei for Hanzi, and Arial and Times New Roman for Latin script. To reflect typical document formatting, text colors are sampled from a set of dark grayscale tones, with a strong bias toward black as detailed in Appendix~\ref{sec:data-specs}. We fix the random seed to 42 for reproducibility. To encourage generalization during fine-tuning, the option format (letters or numbers) is chosen uniformly at random for each synthetic instance.
\begin{table}[t]
\centering
\resizebox{\linewidth}{!}{
\setlength{\tabcolsep}{4pt} 
\begin{tabular}{lcrr}
\toprule
\textbf{Dataset} & \textbf{Languages} & \textbf{Used} & \textbf{Total}\\
\midrule
M3Exam          &  en (610), it (228), zh (351) & 1,189   & 12,317     \\
CMMU            & zh & 1,095   & 3,603      \\
M4U             & de & 2,183   & 8,931      \\
Pinocchio       & it & 1,392   & 136,849   \\
MMMU-Pro        & en & 1,109   & 5,190    \\
\midrule
\textbf{Total} & \textbf{4} & \textbf{6,968} & \textbf{166,890}  \\
\bottomrule
\end{tabular}
}
\caption{Number of questions used from each dataset compared to the total available questions.}
\label{tab:data-coverage}
\end{table}

\begin{table*}[t]
\centering
\begin{adjustbox}{width=\textwidth}
\begin{tabular}{lcccc!{\vrule width 1pt}ccccccccc!{\vrule width 1pt}cc}
\toprule
\textbf{Model} &  \textbf{zh} & \textbf{en} & \textbf{it} & \textbf{de} & \textbf{hr} & \textbf{hu} & \textbf{ar} & \textbf{fr} & \textbf{pl} & \textbf{es} & \textbf{bg} & \textbf{sr} & \textbf{sk} & \textbf{Rel. avg.} & \textbf{Avg.} \\
\midrule
\textit{Train split} & 3308 & 1992 & 2571 & 2573 & 3207 & 3122 & 293 & 199 & 2285 & 190 & 1648 & 887 & -- & -- & -- \\
\textit{Test split} & 600 & 347 & 562 & 279 & 585 & 535 & 517 & 224 & 100 & 100 & 400 & 502 & 46 & -- & -- \\
\midrule
\textbf{Non-FT}   & 24.8 & 21.3 & 23.1 & 29.0 & 25.3 & 27.1 & 23.2 & 34.8 & 22.0 & 31.0 & 30.2 & 22.5 & 17.4 & 24.6 & 25.5\\
\textbf{FT-EV}   & \textbf{32.5} & 22.5 & 32.4 & 42.3 & \textbf{32.3} & 30.3 & 25.0 & \textbf{47.8} & \textbf{30.0} & \textbf{67.0} & \textbf{32.5} & \textbf{27.9} & 37.0 & 32.4 & \textbf{35.3} \\
\textbf{FT-EV+SYN}   & 30.8 & \textbf{23.6} & \textbf{32.6} & \textbf{46.2} & 31.8 & \textbf{31.0} & \textbf{26.7} & 44.2 & 22.0 & 59.0 & 28.2 & 25.7 & \textbf{50.0} & \textbf{33.3} & 34.8\\

\midrule
\rowcolor{gray!20}
\textbf{InternVL3-2B} & 27.8 & 21.3 & 33.8 & 35.5 & 27.4 & 27.1 & 14.3 & 47.8 & 24.0 & 43.0 & 21.3 & 29.9 & 19.6 & 29.6 & 28.7\\
\rowcolor{gray!20}
\textbf{LLaVA-NeXT} & 14.2 & 18.7 & 25.6 & 24.7 & 5.8 & 18.9 & 3.3 & 23.2 & 23.0 & 29.0 & 15.8 & 26.3 & 17.4 & 20.3 & 17.5\\
\bottomrule
\end{tabular}
\end{adjustbox}
\vspace{0.5em}

\caption{Performance comparison of the non-fine-tuned PaliGemma model (\textbf{Non-FT}), fine-tuning on EXAMS-V (\textbf{FT-EV}), and fine-tuning on EXAMS-V with synthetic data (\textbf{FT-EV+SYN}). Results are shown alongside strong vision-language baselines (InternVL3-2B and LLaVA-NeXT-Mistral-7B). \textbf{Rel. avg.} is the average over zh, en, it, and de; \textbf{Avg.} is the overall average across all languages. The number of training and test instances for FT-EV+SYN is shown in the top. Bolded values indicate the best scores within the main results in the top three rows.}
\label{tab:results}
\end{table*}

\section{Experiments}

\subsection{Experimental Setting}

We fine-tune the PaliGemma 2-mix variant with 448$^2$ pixel input resolution. We freeze the vision encoder and the projection layer, training the language decoder for 5 epochs using AdamW with a learning rate of $2 \times 10^{-5}$, weight decay of $1 \times 10^{-6}$, batch size of 64, Eager attention, and a learning rate schedule with linear warm-up over the first 0.05\% of the training steps followed by cosine decay.

To assess the utility of our synthetic text-in-image dataset, we fine-tuned two variants of the model under comparable training settings. The first variant is trained on a combination of the EXAMS-V training split and our synthetic data (FT-EV+SYN), while the second variant was trained exclusively on the original EXAMS-V training split without any synthetic augmentation (FT-EV).

We report results on the EXAMS-V test set, in term of accuracy of the multiple-choice answer that the model generates. As additional strong baselines, we include InternVL3-2B \cite{chen2024expanding} and LlaVA-Next (Mistral-7B) \cite{liu2024llavanext}.

\subsection{Results}

\paragraph{Main Results} 

The main results are reported in Table~\ref{tab:results}, along with the number of train and test data points available for each language in the base dataset, EXAMS-V, as these values become relevant to the discussion below. 

Both of our fine-tuned models, FT-EV and FT-EV+SYN, outperform the off-the-shelf PaliGemma 2 model (Non-FT). FT-EV+SYN achieves the highest average accuracy across the four augmented languages (zh, en, it, de) at 33.3\%, outperforming FT-EV (32.4\%), InternVL3-2B (28.7\%), and LLaVA-NeXT (20.3\%) by 0.9, 4.6, and 13.0 percentage points, respectively. It surpasses FT-EV in 3 out of the 4 languages, with the largest gain in German (+3.9 points). The only exception is Chinese, where performance slightly declines by 1.7 percentage points, possibly because this language is already well-represented in EXAMS-V (with 3308 train data points), reducing the benefit of additional synthetic data.  

On average across all 13 languages, our targeted data augmentation leads to a slight decrease of 0.5 percentage points, possibly due to representational bias toward the augmented subset. 
Interestingly, several non-augmented languages show improvements, suggesting that synthetic data can enhance cross-lingual generalization. For example, Arabic improves by 1.7 points and Hungarian by 0.7 points over FT-EV. Slovak shows an even larger improvement of 13.0 points, but this may be influenced by the small number of test instances in Slovak (only 46), which can increase variance in performance estimates. Other languages also have limited test coverage; for example, Spanish and Polish each have only 100 test instances, which may explain the notable performance drop observed for FT-EV+SYN compared to FT-EV (Spanish: 67.0 to 59.0; Polish: 30.0 to 22.0).

\begin{table}[!h]
\centering
\scalebox{0.9}{
\begin{tabular}{lccccc}
\toprule
\textbf{Model} & \textbf{zh} & \textbf{en} & \textbf{it} & \textbf{de} & \textbf{Avg.}\\
\midrule
\textbf{FT-zh}         & \textbf{32.3}      & 23.3     & 28.1    & 31.5 & 28.8\\  
\textbf{FT-en}         & 29.3     & \textbf{25.4} & 25.3     & 30.8  &    27.7\\
\textbf{FT-it}         & 28.5     & 18.7     & 26.7 & 34.1  &  27.0 \\
\textbf{FT-de}         & 26.5      & 21.6     & 28.8    & 34.1 & 27.8\\
\textbf{FT-EV+SYN} & 30.8 & 23.6 & \textbf{32.6} & \textbf{46.2} & \textbf{33.3} \\

\bottomrule
\end{tabular}
}
\caption{Accuracy of language-specific vs. mixed-language fine-tuned models, evaluated per language. All of these models are trained with augmented data.}
\label{tab:mix-vs-single}
\end{table}

\begin{figure}[t]
    \centering
    \includegraphics[width=\columnwidth]{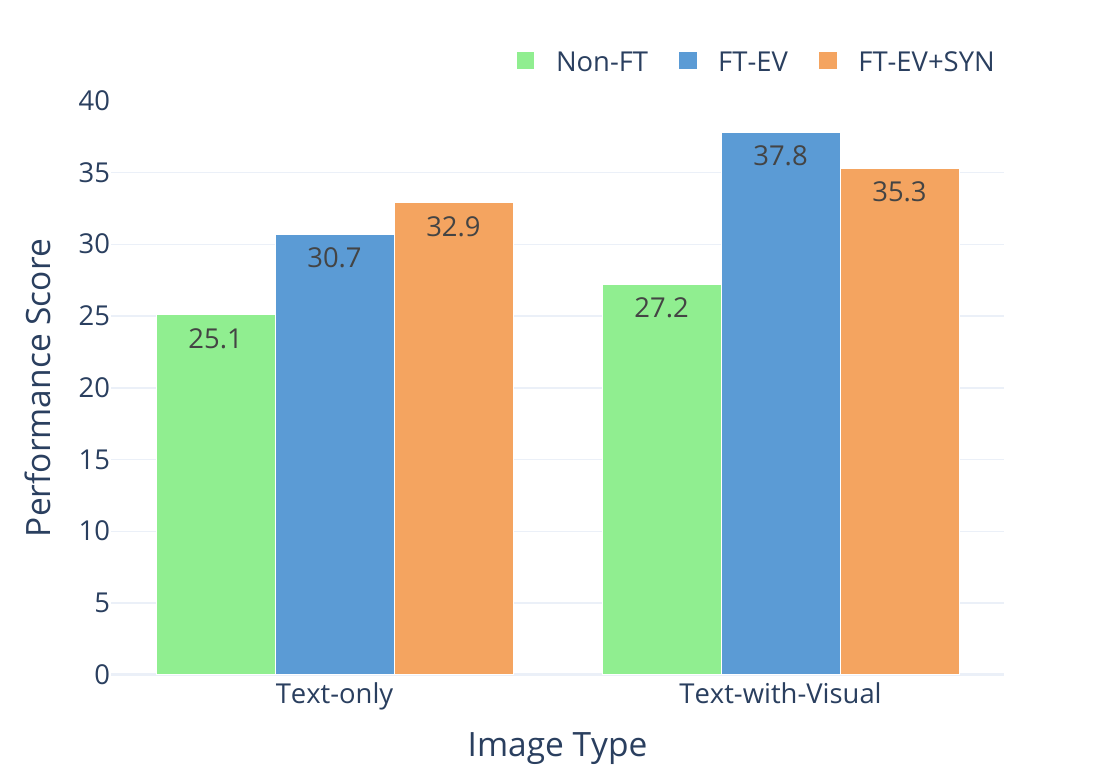}
    \caption{Performance comparison of fine-tuned and non-fine-tuned models across different image types, averaged over the four target languages.}
    \label{fig:image_type comparison.}
\end{figure}

\paragraph{Separate Modality Analysis} 
EXAMS-V includes two image formats: text-only images and images containing both text and visuals (e.g. figures). Here, we investigate how our data augmentation affects each subset. As seen if Figure~\ref{fig:image_type comparison.}, both fine-tuned models, FT-EV and FT-EV+SYN, outperform the non-fine-tuned baseline across both formats. FT-EV+SYN achieves the highest accuracy on text-only images (32.9\%), while FT-EV performs best on text-with-visual images (37.8\%.) This is an unexpected finding, as it suggests that our synthetic data points, designed to contain both text and visuals, benefit text-only questions, but not questions containing both text and visuals. 

Another interesting observation in Figure~\ref{fig:image_type comparison.} is the relatively higher performance of all models on the text-with-visuals portion of the data, compared to text-only. Contrary to prior findings on the relative complexity of multimodal questions, here we see these questions emerging as easier for the models. It remains to be explored whether this is a property of the data or of the models.

\paragraph{Monolingual Training} 
Having established the performance of our model under multilingual training, we now experiment with monolingual training to assess the extent of cross-lingual transfer or interference. As shown in Table~\ref{tab:mix-vs-single}, we fine-tune separate models on monolingual augmented subsets of the dataset and compare their accuracy to that of our multilingual model, FT-EV+SYN. The multilingual model demonstrates superior performance compared to its monolingual counterparts on average across the four languages, as well as in both German and Italian (by 12.1 and 5.9 percentage points, respectively). This considerable gap highlights the benefits of cross-lingual training.

We further observe an intriguing cross-lingual effect: the FT-zh and FT-it outperform the Italian monolingual model, FT-it, on the Italian test set, by a considerable margin of 1.5 to 2 points. This may be attributed to a distributional mismatch between the augmented Italian augmented data and the Italian instances in EXAMS-V.

\section{Conclusion}
In this study, we augment the EXAMS-V dataset with  synthetic text-in-image instances for 4 languages. Our experiments demonstrate improved performance across the four languages on average, albeit on text-only questions and not on questions containing visuals. We find that multilingual fine-tuning outperforms monolingual fine-tuning on average, indicating a positive cross-lingual transfer.

\bibliography{custom}
\appendix
\section{Hyperparameters}
\label{sec:params}

\begin{table}[H]
\centering
\resizebox{\columnwidth}{!}{%
\begin{tabular}{l|l}
\textbf{Parameter}        & \textbf{Value} \\
\hline
Attention Implementation  & Eager\\
Learning Rate             & 2e-5 \\
Weight Decay              & 1e-6 \\
Batch Size                & 64 \\
Optimizer                 & adamw\_torch  \\
Scheduler                 & Warm-up + Cosine Decay \\
Epochs                    & 5 \\
\end{tabular}
}
\caption{Fine-tuning hyperparameters.}
\end{table}

\section{Data Specifications}
\label{sec:data-specs}

\begin{table}[H]
\centering
\small
\resizebox{\columnwidth}{!}{%
\begin{tabular}{lc}
\toprule
\textbf{Script} & \textbf{Fonts} \\
\midrule
Hanzi &
\begin{tabular}[t]{@{}l@{}}
Microsoft YaHei, SimSun, FangSong, \\
SimHei, Alibaba PuHuiTi Regular
\end{tabular} \\
Latin &
\begin{tabular}[t]{@{}l@{}}
Arial, Times New Roman, Courier New, \\
Verdana, Calibri
\end{tabular} \\
\bottomrule
\end{tabular}
}
\caption{Fonts used for generating Hanzi and Latin rendered instances.}
\label{tab:fonts}
\end{table}

\begin{table}[H]
\centering
\small
\resizebox{\columnwidth}{!}{%
\begin{tabular}{lc}
\toprule
\textbf{RGB Value} & \textbf{Sampling Weight (\%)} \\
\midrule
(0, 0, 0) - Black           & 90 \\
(20, 20, 20)              & 2 \\
(43, 43, 43)              & 2 \\
(82, 82, 82)              & 2 \\
(138, 138, 138)           & 2 \\
(168, 168, 168)           & 2 \\
\bottomrule
\end{tabular}
}
\caption{Grayscale RGB values used for text rendering, along with sampling weights.}
\label{tab:gray-colors}
\end{table}

\end{document}